\documentclass[11pt]{article}

\usepackage[margin=1in]{geometry}
\usepackage[T1]{fontenc}
\usepackage{lmodern}
\usepackage{microtype}
\usepackage{amsmath,amssymb}
\usepackage{booktabs}
\usepackage{graphicx}
\usepackage{float}
\usepackage[font=small]{caption}
\usepackage[square,numbers,sort&compress]{natbib}
\usepackage{xcolor}
\usepackage[colorlinks=true,linkcolor=blue,urlcolor=blue,citecolor=blue]{hyperref}

\title{Testing Central Structure for Complex Networks}
\author{%
Qilong Wu\textsuperscript{1},
Yuxiao Li\textsuperscript{2},
Simeng Zhang\textsuperscript{1},
Xinying Liu\textsuperscript{1},
Yan Liu\textsuperscript{3},
Zengyou He\textsuperscript{1,*}\\[0.5em]
\small\textsuperscript{1}School of Software, Dalian University of Technology, China\\
\small\textsuperscript{2}College of Electrical Engineering, Sichuan University, China\\
\small\textsuperscript{3}School of Software Engineering, Dalian University, China\\[0.25em]
\small\textsuperscript{*}Corresponding author: \href{mailto:zyhe@dlut.edu.cn}{zyhe@dlut.edu.cn}
}

\begin{document}

\maketitle

\begin{abstract}
Centrality measures rank nodes even when the observed differences can be
explained by ordinary structural variation. We therefore study a preceding
question: whether a network contains a statistically identifiable central
structure at all. We formulate this question as a network-level hypothesis
testing issue using the adjacency spectral radius as the test statistic and estimate an
empirical $p$-value from randomized networks generated under a specified null
model. The framework is evaluated using $G(N,M)$, Chung--Lu, and two
implementations of the configuration model on simulated benchmarks and 55 real
networks. Controlled experiments further vary the number and connection
weights of designated central nodes while holding network size and density
fixed. The results show a clear negative association between central-structure
strength and the empirical $p$-value. They also demonstrate that rejection decisions
depend substantially on the structural properties preserved by the null
model: the detected network sets are not simply nested across models. 
\end{abstract}


\section{Introduction}
\label{sec:introduction}
Complex networks provide a common language for describing systems whose
elements interact through nontrivial patterns of connectivity, including
social, technological, biological, and transportation systems
\citep{newman2003structure}. Many such
networks exhibit a pronounced central structure. A recurring question is whether a small number of nodes occupy an exceptional
structural position. Such nodes may coordinate
information, connect otherwise separated groups, mediate transport, or act as
particularly effective targets for intervention. The practical consequences of
this question are substantial: influential spreaders are relevant to epidemic
control and information diffusion, while structurally important nodes can be
prioritized for monitoring, immunization, infrastructure protection, or
resource allocation \citep{kitsak2010identification,morone2015influence}.
Accordingly, identifying key nodes has become a fundamental research problem in complex-network
analysis.

A large body of work has developed measures for ranking nodes by different
notions of importance \citep{lu2016vital,ces.2025.34}. Degree centrality favors nodes with many immediate
connections \citep{freeman1978centrality}, betweenness centrality emphasizes shortest-path brokerage \citep{freeman1977set}, and
closeness centrality measures global reachability \citep{sabidussi1966centrality}.
Spectral and prestige-based measures, including eigenvector centrality and
Katz centrality, account for the importance of a node's neighbors
\citep{bonacich1987power,katz1953new}. More recent studies
connect centrality to dynamical influence and seek nodes that maximize the
spread of information or disease \citep{kitsak2010identification,morone2015influence}.
These methods have produced useful and often highly informative rankings.
Nevertheless, they generally presuppose the existence of meaningful
central nodes: they produce a ranking and thereby designate the
highest-ranked nodes as central or influential ones even when the network is nearly
structurally homogeneous, or when the apparent prominence of those nodes
could plausibly arise from ordinary random variation.

This observation points to a prior question that has received
little attention: does the network contain a statistically identifiable
central structure at all? A node can be the
largest-degree node in any finite network without being an exceptional hub in a
statistical sense. In a regular or nearly homogeneous network, a centrality evaluation 
algorithm must still order nodes, but that ordering need not support the claim
that one or several nodes play a special structural role. Hence, the investigation on the existence of central nodes should
precede the selection of a particular centrality measure. Without this
preliminary diagnostic, downstream key-node analysis can report random artifacts where no
single node is meaningfully distinguished from its peers.

An earlier statistical treatment of this question was proposed by
\citet{tallberg2004testing}, who tested graph centralization using group-level
indices derived from degree and closeness centralities, with a Bernoulli graph
as the null model and a centralized blockmodel as the alternative. This work
demonstrated the value of formal centralization testing, but the resulting
procedures depend on the selected node-level centrality measure and on a
restricted blockmodel specification. Their applicability to other forms of
central structure and other background network properties is therefore
limited.

In this work, we formulate the centrality existence testing issue as a
hypothesis-testing problem. For a network with adjacency matrix $A$,
we use its spectral radius, the largest eigenvalue of $A$, as a network-level
statistic for testing whether connectivity is unusually concentrated. We compare the observed
spectral radius with the null distribution generated by a specified random-graph model
and estimate an empirical
$p$-value. A small $p$-value indicates that the observed network has a more
pronounced central structure than expected under that null model.

We evaluate the proposed framework through two complementary experiments. The
first applies the test to synthetic and real networks to examine the robustness
of its conclusions across different random-graph null models and randomized network numbers.
The second uses controlled synthetic networks in which the number of central
nodes and their connection probabilities are varied while the network size and
density are held fixed, allowing us to assess whether the test responds
systematically to changes in the strength of the central structure. In these controlled
experiments, the empirical $p$-values generally decrease as the central
structure becomes stronger, indicating that our method is increasingly likely
to classify the observed network as containing central nodes.

This study makes three contributions. First, it places the question of whether
a network contains a statistically identifiable central structure before the
usual task of ranking individual nodes.
Second, to the best of our knowledge, it introduces the first hypothesis-testing
framework for determining whether a network contains central nodes. Third, it provides a controlled evaluation on synthetic networks and an empirical
evaluation on real networks.

\section{Method}
\label{sec:method}

\subsection{Problem statement}
\label{sec:problem_statement}
Let $G = (V,E)$ be an unweighted and undirected graph, where $V = \{v_1, v_2, \dots, v_N\}$
is a set of $N$ nodes, $E = \{e_1, e_2, \dots, e_M\}$ is a set of $M$ edges,
and $A=(a_{ij})\in\{0,1\}^{N\times N}$ is its corresponding adjacency matrix.
We aim to determine whether a network contains a statistically identifiable
central structure before the usual task of ranking individual
nodes. Here, a central structure means that the network's connectivity is
unusually concentrated around one or more dominant nodes.
The null hypothesis and the alternative hypothesis can be stated as:
\begin{equation*}
\left\{
\begin{aligned}
H_0 &: \text{the central structure is not pronounced relative to the specified null model},\\
H_1 &: \text{the central structure is pronounced relative to the specified null model}.
\end{aligned}
\right.
\end{equation*}

To quantify whether we should reject the null hypothesis,
we need to first choose a test statistic.
Because the network is undirected and contains no self-loops, $A$ is symmetric
and has a zero diagonal. 
We use the spectral radius:
\begin{equation}
    T(G)=\rho(A)=\lambda_{\max}(A),
    \label{eq:test_statistic}
\end{equation}
as a network-level statistic for testing whether connectivity is unusually
concentrated. We compute the spectral radius using the Lanczos algorithm
\citep{lanczos1950iteration}. For
a sparse graph, its computational complexity is approximately $O(qM)$, where
$q$ is the number of iterations and is typically much smaller than $M$. For a
nonnegative symmetric adjacency matrix, the spectral
radius is the maximum absolute value of its eigenvalues. If $d_{\max}$ and $\bar d=2M/N$ denote the
maximum and average degrees, respectively, then:
\begin{equation}
    \max\{\bar d,\sqrt{d_{\max}}\}
    \leq \rho(A) \leq d_{\max},
    \label{eq:spectral_bounds}
\end{equation}
for every nonempty simple undirected network
\citep{chung1997spectral,cvetkovic2010introduction}. Thus, an unusually
high-degree node necessarily raises the scale of the spectral radius, while
the leading eigenvalue also incorporates how the node and its neighbors are
embedded in the rest of the network. The statistic can therefore respond to
both a single dominant node and a group of mutually reinforced central nodes.

The spectral radius also has a well-established dynamical interpretation. It
characterizes the network's overall connectivity strength and is closely
related to spreading potential. In spectral analyses of the
Susceptible--Infected--Susceptible (SIS) model, the epidemic threshold is governed by the
inverse of the adjacency spectral radius. A larger spectral radius therefore
corresponds to a lower epidemic threshold and a greater potential for
large-scale spreading \citep{chakrabarti2008epidemic}. These structural and
dynamical properties motivate our choice of $\rho(A)$ as the test statistic.


\subsection{Random graph models}
\label{sec:random_graph_models}

The null model specifies which properties of the observed network are treated
as ordinary background structure. We consider three null-model families with
different levels of constraint. The configuration family is evaluated through
two construction methods described below.

\subsubsection{E-R model}

The Erd\H{o}s--R\'enyi (E-R) model is one of the simplest and most widely used
random graph models. Two closely related variants are commonly considered.
The $G(N,M)$ model samples uniformly from the collection of all simple graphs
with $N$ nodes and exactly $M$ edges \citep{erdos1959}. By contrast, the
$G(N,p)$ model considers every unordered pair of distinct nodes independently
and places an edge between them with probability $p$ \citep{gilbert1959}. In
both variants, the generated network is a simple graph: self-loops and
multiple edges are excluded.

In this study, we adopt the $G(N,M)$ model. This allows the test to compare the
observed network with randomized networks of the same size and density.
Each possible set of $M$ edges is selected with probability:
\begin{equation}
    \Pr(G^*=g)=\binom{\binom{N}{2}}{M}^{-1}.
    \label{eq:gnm}
\end{equation}

The model therefore preserves the average degree, but it does not preserve the
observed degree sequence. It asks whether the observed network has a more
pronounced central structure than a random network with the same global size
and density. Rejection under this model may
therefore be driven by exceptional degree heterogeneity, by an unusual wiring
pattern, or by both.

\subsubsection{Configuration model}

The configuration model preserves the observed degree sequence in the generated
random networks \citep{newman2001random}. We consider two construction methods of this model.
In the standard stub-matching method, each node is assigned a number of stubs
equal to its observed degree, and the stubs are paired at random. This
procedure may generate self-loops and parallel edges. To obtain a simple
graph, self-loops are removed and parallel edges are merged; as a result, the
observed degree sequence is preserved only approximately.
Alternatively, an exact-degree simple graph can be obtained by randomizing
the observed graph through degree-preserving double-edge swaps. A valid swap
replaces two edges $\{u,v\}$ and $\{x,y\}$ with $\{u,y\}$ and
$\{x,v\}$, provided that no self-loop or parallel edge is created. Repeating
such swaps changes the wiring pattern while preserving every node degree
exactly \citep{fosdick2018}.

By controlling for degree sequence, the configuration model provides a more
realistic null model than $G(N,M)$. Rejecting the degree-preserving null
does not merely indicate that the network contains high-degree nodes. Rather,
it indicates that, even among randomized networks with the same degree
sequence, the observed network has a more pronounced central structure.

\subsubsection{Chung--Lu model}

The Chung--Lu model preserves a specified degree sequence in expectation
\citep{chung2002connected,chung2003spectra}. In this model, the observed
degree sequence $(d_1,d_2,\ldots,d_N)$ is used as the target expected degree
sequence. Distinct nodes $v_i$ and $v_j$ are connected independently with
probability:
\begin{equation}
    p_{ij}=\min\left\{\frac{d_i d_j}{\sum_{k=1}^{N}d_k},1\right\},
    \qquad i\neq j.
    \label{eq:chung_lu}
\end{equation}

Unlike the configuration model, which fixes the degree of every node, the
Chung--Lu model allows the realized degrees to fluctuate around their expected
values. It therefore imposes a less restrictive constraint than the
configuration model. The use of such a null model tests whether, after controlling
for the expected degree sequence, the observed network still exhibits a more
pronounced central structure than the corresponding randomized networks.

\subsection{\texorpdfstring{$\boldsymbol{p}$}{p}-Value calculation}
\label{sec:hypothesis_test}

For a selected null model, we first compute the observed statistic
$T_{\mathrm{obs}}=T(G)=\rho(A)$. We then independently generate $B$
randomized networks $G^{(1)},\ldots,G^{(B)}$ under that null model and compute
their spectral radii:
\begin{equation}
    T_b=T\!\left(G^{(b)}\right), \qquad b=1,\ldots,B.
\end{equation}
These values form an empirical null distribution of the test statistic.

\begin{figure}[!t]
    \centering
    \makebox[\textwidth][c]{%
        \includegraphics[width=0.95\textwidth]{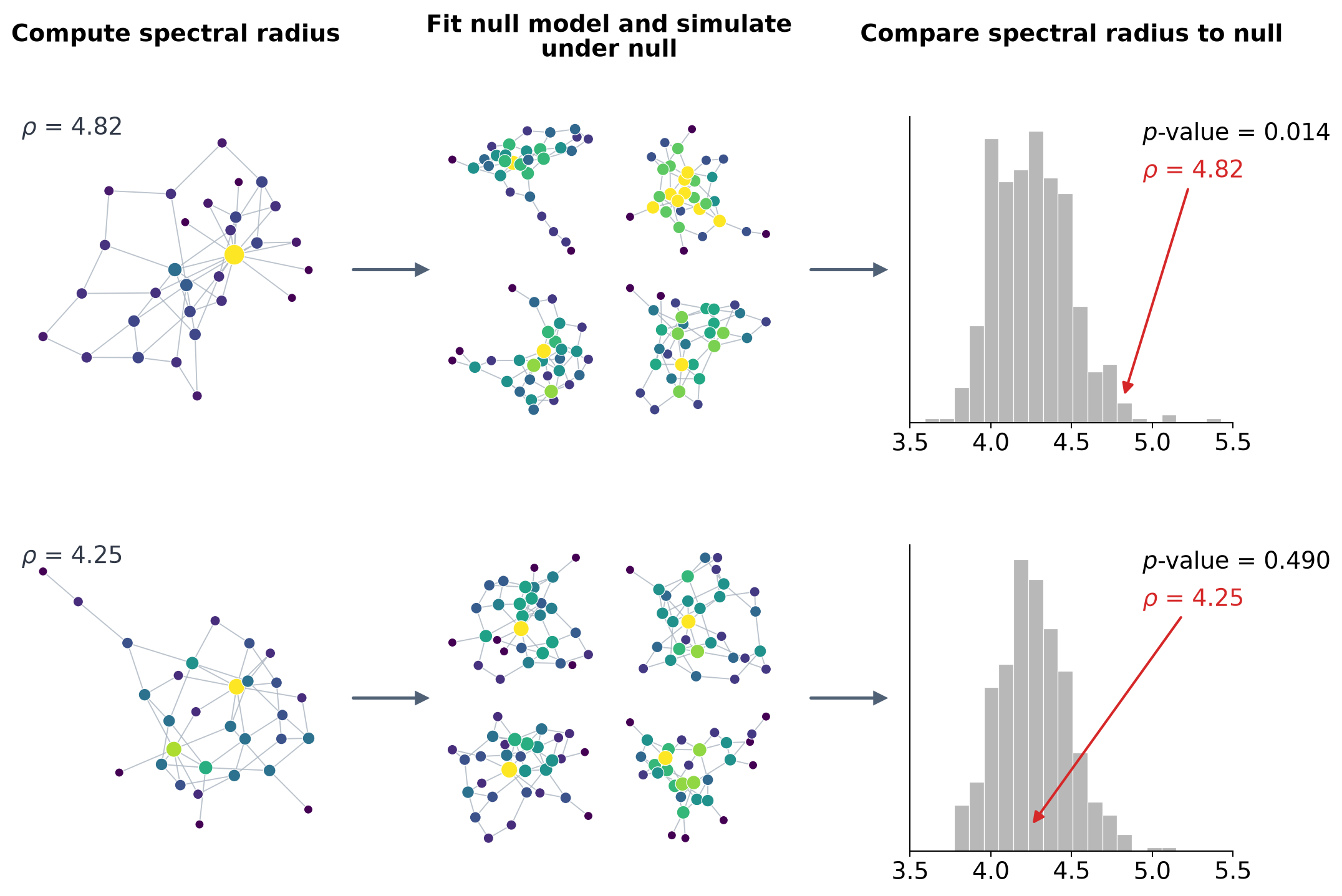}%
    }
    \caption{An illustration of our method with
    two examples: a network with a pronounced central structure (top) and a
    network consistent with the null model (bottom). We first
    compute the spectral radius of the observed network and construct an
    empirical null distribution from the spectral radii of multiple networks
    simulated under the $G(N,M)$ model. We then compare the observed
    spectral radius with the resulting empirical null distribution to decide
    whether to reject the null hypothesis.}
    \label{fig:method_overview}
\end{figure}

Because a pronounced central structure is expected to produce a large
spectral radius, we conduct a one-sided test. The empirical
$p$-value is calculated as:
\begin{equation}
    p\text{-value}
    =\frac{1}{B}\sum_{b=1}^{B}
    \mathbb{I}\!\left[
        T\!\left(G^{(b)}\right)\geq T(G)
    \right],
    \label{eq:p_value}
\end{equation}
where $\mathbb{I}[\cdot]$ is the indicator function. Equivalently,
$p$-value is the proportion of randomized networks whose spectral radius is
at least as large as that of the observed network. The significance testing
procedure is illustrated in
Fig.~{\hypersetup{linkcolor=blue}\ref{fig:method_overview}}.

Given a significance level $\alpha$, we reject $H_0$ when
$p\text{-value}<\alpha$. A small $p$-value indicates that the observed spectral
radius lies in the upper tail of the null distribution and that the network
has a more pronounced central structure than expected.
If $H_0$ is not rejected, the result means that the available
evidence is insufficient to distinguish the observed central structure from
random variation under that null model. The test concerns the network as a whole and does not assess the
centrality of any particular node.

\section{Experiments}
\label{sec:experiments}

We conduct two experiments to evaluate the proposed central-structure test.
The first experiment studies the behavior of the test on simulated and real
networks while varying the number of random networks and the random-graph null
model. The second uses controlled synthetic networks in
which the number of central nodes and their connection probabilities are varied
while the network size and density are held fixed, allowing us to assess whether
the test responds systematically to changes in the strength of the central
structure.
Unless stated otherwise, all tests use a significance level of
$\alpha=0.05$ and the one-sided empirical $p$-value defined in
Eq.~\eqref{eq:p_value}.

\subsection{Evaluation on simulated and real networks}
\label{sec:simulation_real_evaluation}

Both the simulated-network and real-network evaluations use
$B\in\{100,500,1000\}$ randomized networks under four null-model
implementations from three model families: $G(N,M)$, Chung--Lu, and two
implementations of the configuration model (configuration-base and
configuration-swap). For each network, we
calculate the spectral radius of the observed network and compute the
empirical $p$-value by comparing it with the empirical null distribution
generated under the corresponding null model.

We use ``configuration-base'' to denote the standard stub-matching
configuration model followed by removal of self-loops and merging of parallel
edges to obtain a simple graph. We use ``configuration-swap'' to denote
degree-preserving double-edge swaps applied to the observed simple graph.
The former only approximately preserves the degree sequence and can change
the edge count after simplification, whereas the latter preserves every node
degree and the total number of edges.

\subsubsection{Simulated networks}
\label{sec:simulated_networks}

We first construct a benchmark with known structural labels. The positive
class contains networks with an explicitly generated central structure,
including star networks, hub--Erd\H{o}s--R\'enyi networks, two-hub networks, and
Barab\'asi--Albert networks with an additional dominant node. The negative class
contains Erd\H{o}s--R\'enyi networks and regular networks.
The benchmark contains 20 positive and 20 negative networks,
with network sizes and densities varied
across the instances.
Because the simulated benchmark has known labels, we report the
true-positive, false-positive, true-negative, and false-negative counts and
the resulting precision, recall, and F1 score in addition to the empirical
$p$-values.



Table~\ref{tab:simulated_results} reports the identification performance on
the simulated benchmark. Under the $G(N,M)$ null model, the identification
results give an F1 score of approximately $0.98$. Under the Chung--Lu null
model, $H_0$ is rejected for
12 of the 20 positive-labeled networks and for none of the 20
negative-labeled networks, giving an F1 score of $0.75$. The two
configuration implementations behave differently. The configuration-base
construction rejects $H_0$ for all 20 positive-labeled networks and all 20
negative-labeled networks.

\begin{table}[htbp]
\centering
\caption{Identification results on the simulated networks. TP, FP,
TN, and FN denote the number of true positives, false positives, true negatives, and false
negatives, respectively. A dash indicates
that the F1 score is undefined because both precision and recall are zeros.}
\label{tab:simulated_results}
\setlength{\tabcolsep}{4pt}
\begin{tabular}{l*{9}{c}}
\toprule
\shortstack[l]{Null model} & $B$ &
TP & FP & TN & FN &
Precision & Recall & \shortstack{F1\\score} & \shortstack{Runtime\\(s)} \\
\midrule
$G(N,M)$                     & 100  & 20 & 1  & 19 & 0  & 0.952 & 1.000 & 0.976 & 5.31 \\
$G(N,M)$                     & 500  & 20 & 1  & 19 & 0  & 0.952 & 1.000 & 0.976 & 26.04 \\
$G(N,M)$                     & 1000 & 20 & 1  & 19 & 0  & 0.952 & 1.000 & 0.976 & 52.20 \\
Configuration-base & 100  & 20 & 20 & 0  & 0  & 0.500 & 1.000 & 0.667 & 15.32 \\
Configuration-base & 500  & 20 & 20 & 0  & 0  & 0.500 & 1.000 & 0.667 & 87.48 \\
Configuration-base & 1000 & 20 & 20 & 0  & 0  & 0.500 & 1.000 & 0.667 & 173.67 \\
Configuration-swap & 100  & 1  & 1  & 19 & 19 & 0.500 & 0.050 & 0.091 & 40.90 \\
Configuration-swap & 500  & 0  & 1  & 19 & 20 & 0.000 & 0.000 & --    & 213.03 \\
Configuration-swap & 1000 & 1  & 1  & 19 & 19 & 0.500 & 0.050 & 0.091 & 421.77 \\
Chung--Lu                      & 100  & 12 & 0  & 20 & 8  & 1.000 & 0.600 & 0.750 & 11.00 \\
Chung--Lu                      & 500  & 12 & 0  & 20 & 8  & 1.000 & 0.600 & 0.750 & 63.47 \\
Chung--Lu                      & 1000 & 12 & 0  & 20 & 8  & 1.000 & 0.600 & 0.750 & 130.78 \\
\bottomrule
\end{tabular}%
\end{table}

These numbers should not be interpreted as a direct performance ranking across
null models, because the models define different null hypotheses and preserve
different network properties. In particular, the simulated positive examples
are constructed primarily by increasing the degrees of designated hub nodes.
The configuration-swap null model preserves those degrees exactly, so the
main signal used to define the positive class is conditioned on rather than
tested. Its low recall is therefore expected for this benchmark: it tests for
additional topology beyond the degree sequence, not merely for the presence of
a high-degree node. The configuration-base construction is not an exact
degree-preserving simple-graph null model; removing self-loops and merging parallel
edges can reduce the null graph's edge count and spectral radius, which helps
explain its large number of false positives. The Chung--Lu model preserves the
degree sequence only in expectation. For highly heterogeneous sequences, simple-graph constraints
can make the realized hub degrees, edge count, and spectral radius lower than observed in some
realizations \citep{chung2002connected,chung2003spectra}. This network-dependent effect may reduce
the empirical $p$-value, but is not universal. Thus, a
rejection under $G(N,M)$, Chung--Lu, or either configuration construction
answers a different structural question and should be interpreted relative to
the corresponding null model.

Moreover, the identification results on this benchmark are largely insensitive
to the number of randomized networks $B$; increasing $B$ mainly increases the
computational cost, with the runtime growing approximately linearly.

\subsubsection{Real networks}
\label{sec:real_networks}

We next apply the test to 55 real networks compiled from the 
Pajek Datasets and the Network Data Repository.
The collection covers networks from social, biological,
technological, transportation, and trade-related settings. Because the data
collection does not provide an independent ground-truth
label for the existence of a statistically identifiable central structure, the
real-network experiment is not treated as a supervised learning task.
Instead, we record the $p$-value and the reject/non-reject decision for each
network, and examine the stability of these conclusions across different
random-graph null models and randomized network numbers.

Table~\ref{tab:real_detection_results} summarizes the number of real networks
for which the null hypothesis is rejected. These rejection counts should not
be interpreted as a universal accuracy ranking. As discussed for the
simulated benchmark, the null models condition on different background
structures, and the configuration-base null model is additionally affected
by the conversion from a multigraph to a simple graph. Because the real
networks have no independent ground-truth labels, these counts cannot be
classified as true or false positives; they should instead be read as
model-dependent rejection frequencies. If all 55 real networks are treated as
positive cases for descriptive comparison, the rejection frequency is
numerically equivalent to recall, although it is not recall validated against
independent labels. Under this convention, the configuration-swap null model shows a much
higher apparent recall on the real networks than on the simulated benchmark.
This difference suggests that real networks contain higher-order organization
beyond the degree sequence, which remains detectable after the degrees are
held fixed.

\begin{table}[!t]
    \centering
    \caption{Real-network rejection results for different null models and
    numbers of randomized networks. Each entry reports the number of rejected
    networks followed by the corresponding percentage of the 55 networks.}
    \label{tab:real_detection_results}
    \begin{tabular}{lccc}
        \toprule
        Null model & $B=100$ & $B=500$ & $B=1000$ \\
        \midrule
        $G(N,M)$ & 53 (96.4\%) & 52 (94.5\%) & 52 (94.5\%) \\
        Configuration-base & 50 (90.9\%) & 51 (92.7\%) & 51 (92.7\%) \\
        Configuration-swap & 30 (54.5\%) & 31 (56.4\%) & 31 (56.4\%) \\
        Chung--Lu & 20 (36.4\%) & 19 (34.5\%) & 19 (34.5\%) \\
        \bottomrule
    \end{tabular}
\end{table}

\begin{figure}[!t]
    \centering
    \includegraphics[width=0.88\textwidth]{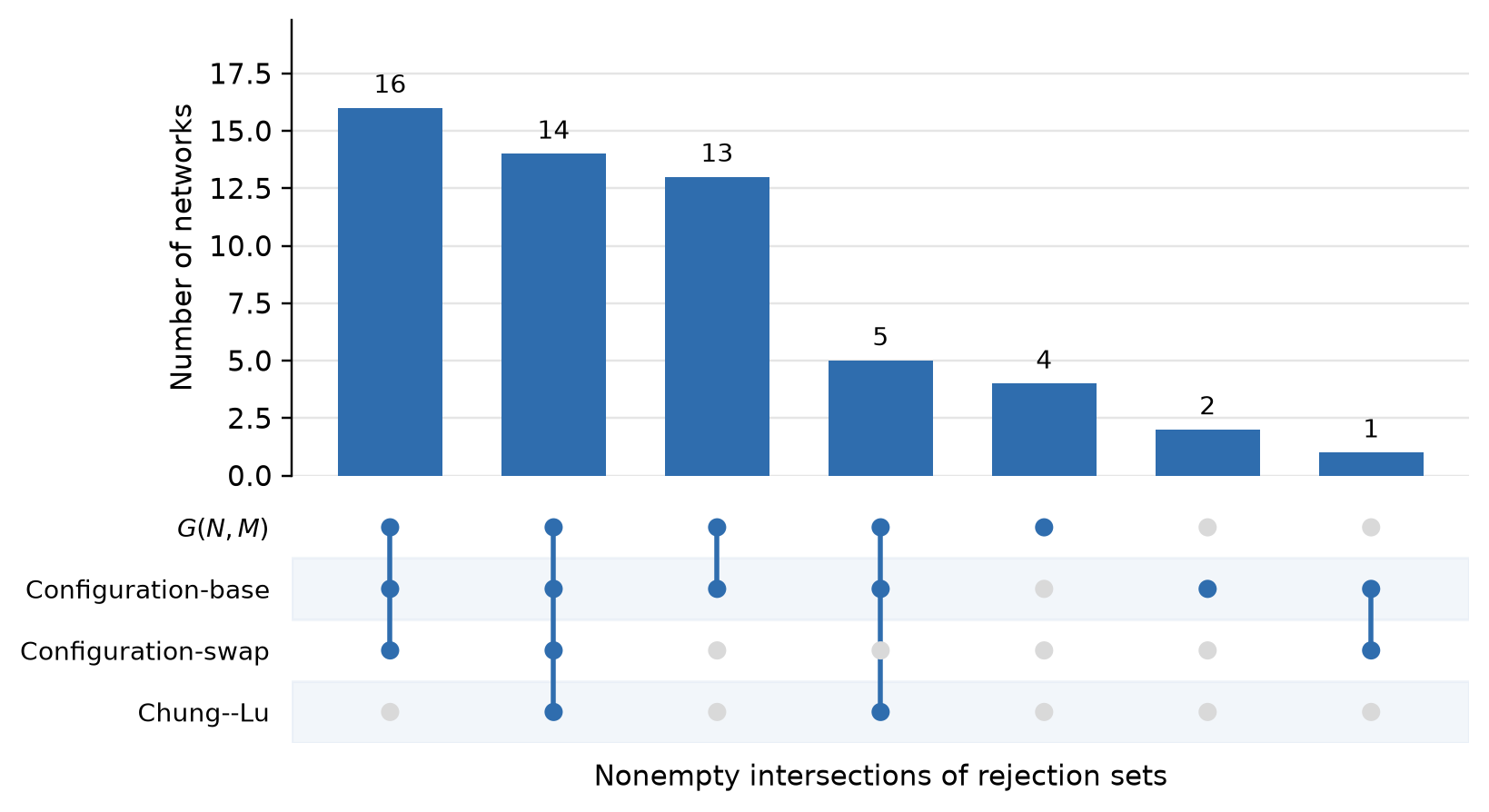}
    \caption{Intersections of the real-network rejection sets at $B=1000$.
    Each bar gives the number of networks rejected by the combination of null
    models indicated by the connected blue dots below; gray dots indicate that
    the corresponding null model does not reject. Only nonempty intersections
    are shown.}
    \label{fig:real_null_model_detection}
\end{figure}

The sets of real networks rejected by the different null models are not
simply nested: rejection under one null model does not imply rejection under
another. Some networks are rejected only after conditioning on the exact
degree sequence, whereas others are rejected under $G(N,M)$ or Chung--Lu but
not under configuration-swap. This pattern, visualized in
Fig.~\ref{fig:real_null_model_detection}, shows that the detected set depends
substantially on which structural properties are treated as ordinary
background variation.

\subsection{Evaluation of central-structure strength}
\label{sec:central_structure_strength}

The second experiment examines how the test responds to changes in the
strength of a central structure. For each fixed network size, edge count, and
connection-weight level, we increase the number of designated central nodes
and track the resulting change in the $p$-value. We repeat
this analysis for several fixed weight levels. We use two fixed-size settings,
$(N,M)=(50,100)$ and
$(N,M)=(100,200)$. The null model is $G(N,M)$. In each generated graph, the first $k$
nodes are designated as central ones, with weight $w$, while all other nodes have
weight one. An unordered pair $(v_i,v_j)$ is sampled without replacement with
probability proportional to the product of its endpoint weights. We vary
\[
 k\in\{1,2,\ldots,\lfloor 0.3N\rfloor\}, \qquad
 w\in\{1.5,2,3,4,6\},
\]
and repeat each condition for 50 random seeds. For each graph, we generate
$B=100$ $G(N,M)$ null networks to calculate the empirical $p$-value.

\begin{figure}[!t]
    \centering
    \includegraphics[width=0.49\textwidth]{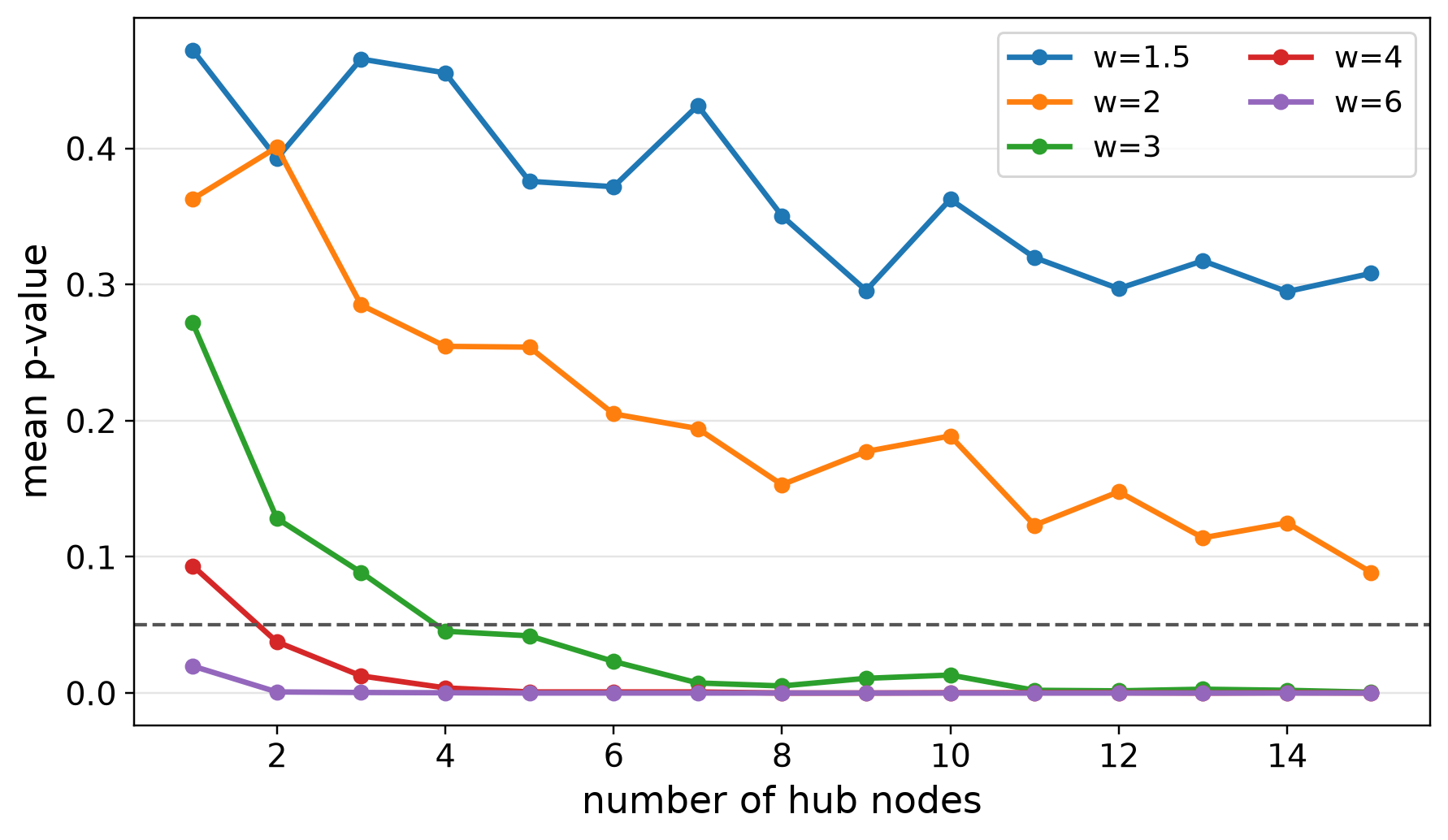}\hfill
    \includegraphics[width=0.49\textwidth]{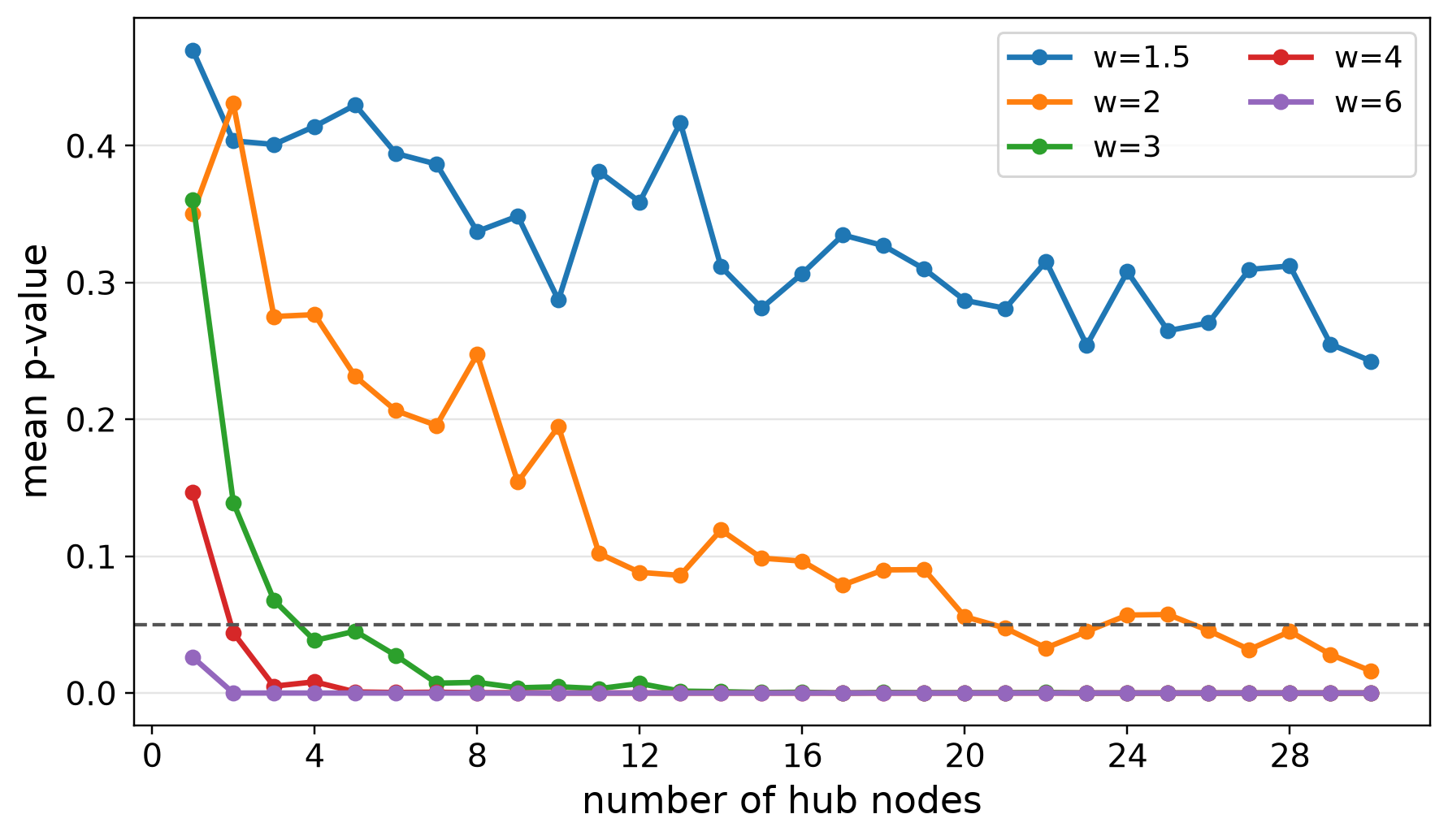}
    \caption{Mean $p$-values as the number of designated
    central nodes increases under the $G(N,M)$ null model. The left panel uses
    $(N,M)=(50,100)$ and the right panel uses $(N,M)=(100,200)$. Each curve
    corresponds to a fixed central-node weight $w$ and averages 50 random
    seeds; the dashed horizontal line indicates $\alpha=0.05$.}
    \label{fig:central_strength_pvalues}
\end{figure}

Fig.~\ref{fig:central_strength_pvalues} shows the mean empirical
$p$-values as the number of central nodes increases for the two network-size
settings. Table~\ref{tab:central_strength_spearman} reports the Spearman rank
correlations between the number of central nodes and the corresponding mean
$p$-values. The experimental results show that increasing the number of
central nodes generally lowers the $p$-value. For $w\geq 3$, however, many of
the mean $p$-values are exactly zero, which compresses the rank variation and
can make the observed Spearman coefficients smaller in absolute value.

\begin{table}[!t]
    \centering
    \caption{Spearman correlations between the number of central nodes and the
    mean $p$-value for the two fixed-size settings.}
    \label{tab:central_strength_spearman}
    \begin{tabular}{lcc}
        \toprule
        $w$ & \multicolumn{2}{c}{Spearman correlations} \\
        \cmidrule(lr){2-3}
        & $(N,M)=(50,100)$ & $(N,M)=(100,200)$ \\
        \midrule
        1.5 & $-0.879$ & $-0.825$ \\
        2.0 & $-0.964$ & $-0.957$ \\
        3.0 & $-0.945$ & $-0.960$ \\
        4.0 & $-0.847$ & $-0.810$ \\
        6.0 & $-0.779$ & $-0.311$ \\
        \bottomrule
    \end{tabular}
\end{table}

\begin{figure}[htbp]
    \centering
    \includegraphics[width=0.95\textwidth]{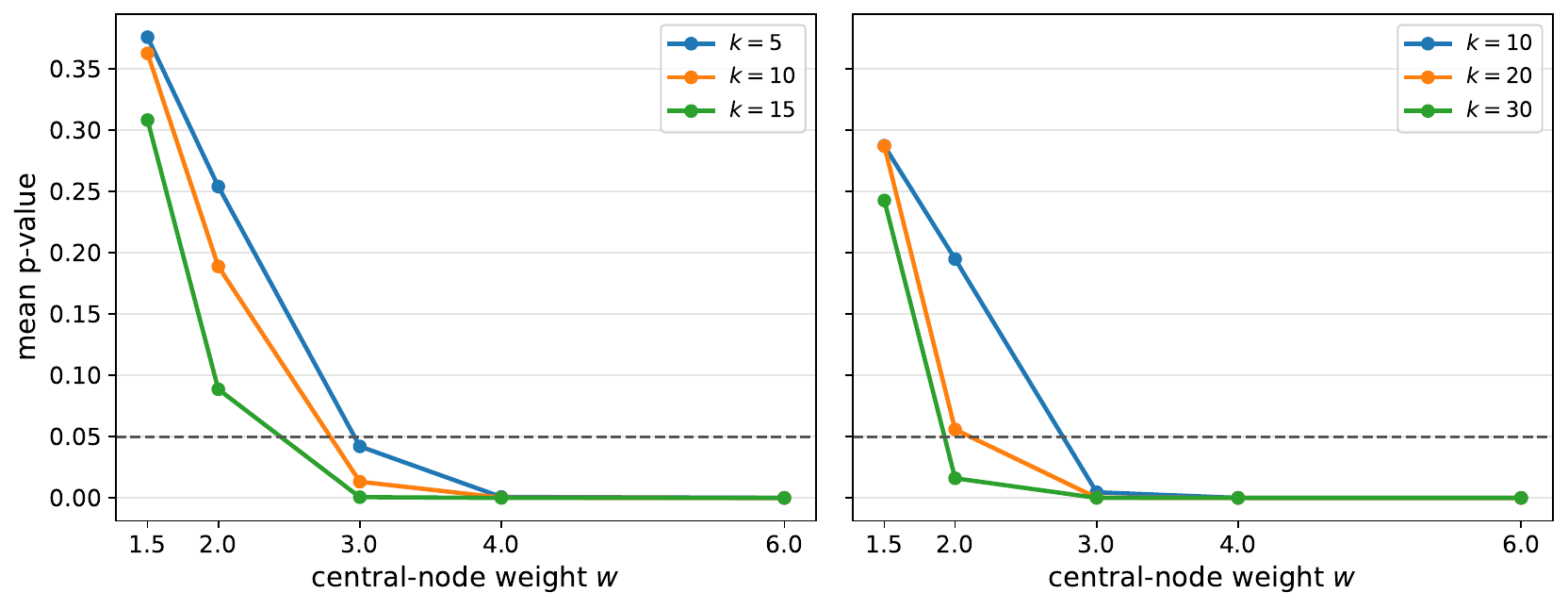}
    \caption{Mean $p$-values as the central-node weight $w$
    increases while the number of central nodes $k$ is held fixed. The left
    panel uses $(N,M)=(50,100)$ with $k\in\{5,10,15\}$, and the right panel
    uses $(N,M)=(100,200)$ with $k\in\{10,20,30\}$. Each point averages 50
    random seeds; the dashed horizontal line indicates $\alpha=0.05$.}
    \label{fig:central_weight_fixed_k}
\end{figure}

Fig.~\ref{fig:central_weight_fixed_k} complements
Fig.~\ref{fig:central_strength_pvalues} by holding the number of central nodes
fixed and varying only their connection weight. In both network-size settings
and for every fixed value of $k$ shown, increasing $w$ generally lowers the
mean $p$-value. 
Taken together, the two figures show that the test responds to both the number
of central nodes and the strength of their preferential connectivity.

\section{Conclusion}
\label{sec:conclusion}

This work introduced a network-level hypothesis test for determining whether a
network contains a statistically identifiable central structure. The test uses
the spectral radius as a measure of connectivity concentration and compares the
observed value with an empirical null distribution generated under a specified
random-graph model. It is intended as a diagnostic before applying node-level
centrality rankings: a rejection indicates that the observed concentration is
unusual relative to the chosen null model.

The experiments show that the test responds systematically to the strength of
the planted structure: increasing either the number of central nodes or their
connection weight generally lowers the empirical $p$-value. They also show
that detection depends substantially on the null-model choices. 

The current study is limited to unweighted, undirected networks, and
the real-network collection does not provide independent labels for the
presence of central structure. Future work could extend the framework to weighted,
directed, and temporal networks, investigate alternative network-level
statistics and more tightly constrained null models, and evaluate the test on
real networks with external structural labels.

\section*{Author Contributions}
Qilong Wu: Writing--review \& editing, Writing--original draft, Visualization, Validation, Resources, Methodology, Formal analysis.
Yuxiao Li: Writing--review \& editing, Writing--original draft, Formal analysis.
Simeng Zhang: Writing--review \& editing, Validation.
Xinying Liu: Writing--review \& editing, Methodology.
Yan Liu: Writing--review \& editing, Visualization, Formal analysis.
Zengyou He: Writing--review \& editing, Writing--original draft, Supervision, Conceptualization.

\section*{Acknowledgments}
This work has been supported by the Natural Science Foundation of China under Grant No. 62472064,
and Chronic Disease Management Research Project of National Health Commission Capacity Building and
Continuing Education Center (No. GWJJMB202510022150).
\section*{Conflicts of Interest}
The authors declare no conflicts of interest.

\section*{Data Availability Statement}
The data that support the findings of this study are available at Pajek
Datasets and Network Data Repository.
The original Pajek Datasets website
(\url{http://vlado.fmf.uni-lj.si/pub/networks/data/}) is unavailable; these
networks were obtained from its GitHub mirror
(\url{https://github.com/bavla/Nets/tree/master/data/Pajek}).

\bibliographystyle{unsrtnat}
\bibliography{references}

\end{document}